\pdfoutput=1

\documentclass[11pt]{article}

\usepackage{EMNLP2022}

\usepackage{times}
\usepackage{latexsym}

\usepackage[T1]{fontenc}

\usepackage[utf8]{inputenc}

\usepackage{microtype}

\usepackage{inconsolata}

\usepackage{graphicx}
\usepackage{subfigure}
\graphicspath{{img/}}
\usepackage{multirow}
\usepackage{booktabs}

\usepackage{amsmath}

\title{  Causal Inference for Chatting Handoff \thanks{Work in progress}}

\author{Shanshan Zhong, Jinghui Qin, Zhongzhan Huang, Daifeng Li\thanks{Corresponding authour}\\
School of Computer Science and Engineering \\ Sun Yat-sen University
}

\begin{document}
\maketitle
\begin{abstract}

Aiming to ensure chatbot quality by predicting chatbot failure and enabling human-agent collaboration, Machine-Human Chatting Handoff (MHCH) has attracted lots of attention from both industry and academia in recent years. 
However, most existing methods mainly focus on the dialogue context or assist with global satisfaction prediction based on multi-task learning, which ignore the grounded relationships among the causal variables, like the user state and labor cost. These variables are significantly associated with handoff decisions, resulting in prediction bias and cost increasement. 
Therefore, we propose Causal-Enhance Module (CEM) by establishing the causal graph of MHCH based on these two variables, which is a simple yet effective module and can be easy to plug into the existing MHCH methods. 
For the impact of users, we use the user state to correct the prediction bias according to the causal relationship of multi-task. For the labor cost, we train an auxiliary cost simulator to calculate unbiased labor cost through counterfactual learning so that a model becomes cost-aware.
Extensive experiments conducted on four real-world benchmarks demonstrate the effectiveness of CEM in generally improving the performance of existing MHCH methods without any elaborated model crafting.

\end{abstract}

\section{Introduction}
\begin{figure}[t]
  \centering
  \includegraphics[width=0.88\linewidth]{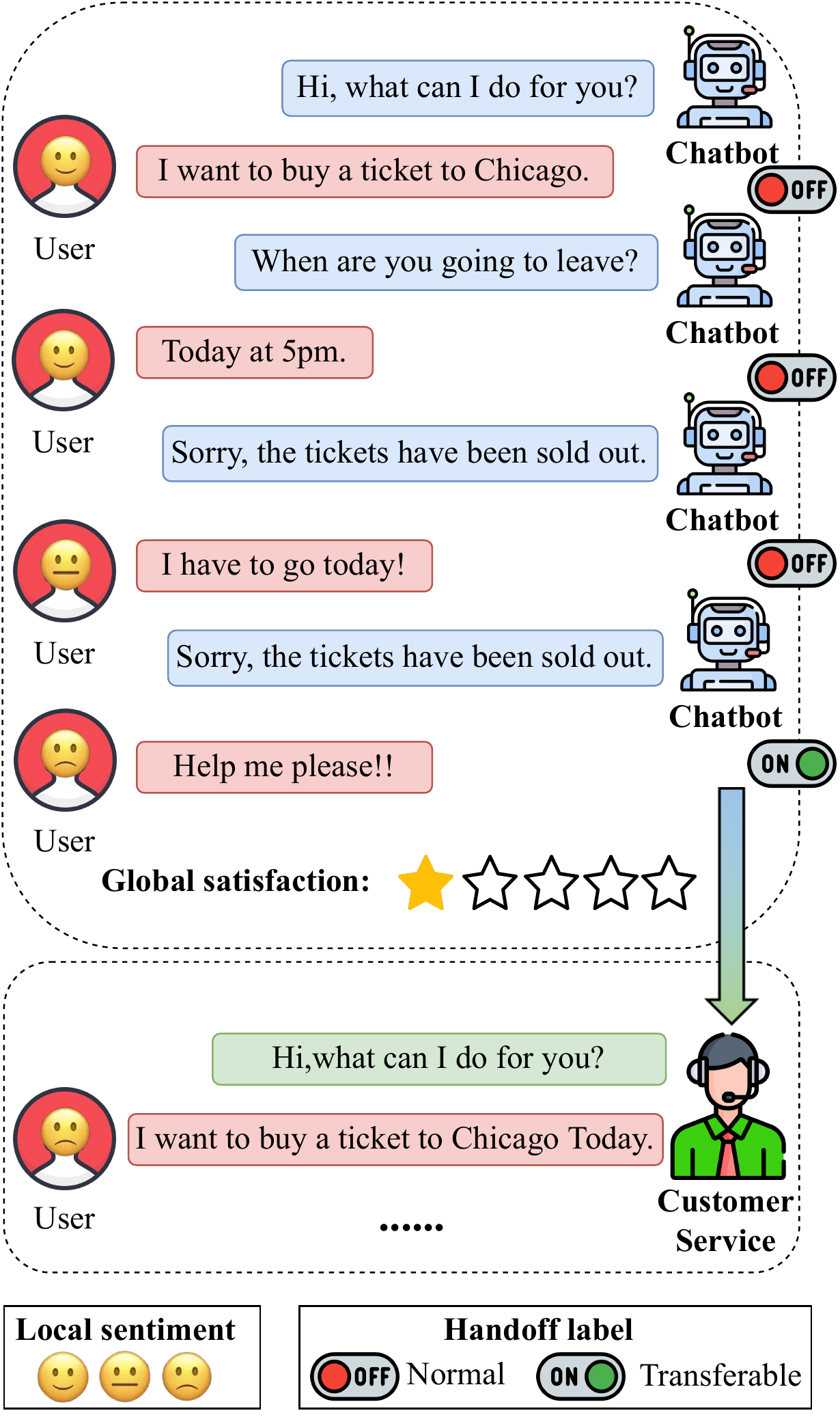}
  \caption{An example of MHCH. Handoff label includes two types "normal" \& "transferable", which denotes whether the chatbot should be transferred to human service. }
\label{fig:dialog}
\end{figure}


In recent years, with the rapid development of deep learning \cite{He_2016_CVPR,ren2015faster}, more and more service-oriented organizations have deployed chatbots to alleviate the problem of limited service resources. Although these chatbots can respond in real-time and save labor cost, they suffer from inappropriate responses and invalid conversations due to the limited quantity of available high-quality training data and the inherent biases \cite{xu2019frequency,liang2022stiffnessaware} of neural networks.
Moreover, the human utterances sometimes are elusive since they are rich in acronyms, slang words, and even content without logic or grammar, which are too obscure for a chatbot to comprehend. 
To alleviate these drawbacks, researchers introduced a human-agent collaboration mechanism named Machine-Human Chatting Handoff (MHCH) to allow a human to take over the dialogue while a robot agent feel confused so that a dialogue can be continued to avoid a bad user experience and reduce the risk of customer churn \cite{liu2021time, liu2021role}.
As shown in Fig.\ref{fig:dialog}, when a chatbot try to address the user's needs by giving an inappropriate response, 
the user will feel disappointed, and give a low global satisfaction score for current dialogue, which means a service failure and may lead to customer loss. 
If deploying with MHCH mechanism, a human can take over the dialogue and give a satisfactory response to meet the user's needs, thus ensuring the user experience and service quality \cite{2017Evaluating}.

In fact, a high-quality MHCH service should consider multiple factors, such as dialogue context, local sentiments, global satisfaction, user state, and labor cost, etc. However, most existing MHCH methods mainly concerned on the dialogue context \cite{liu2021time} or assisting with global satisfaction prediction under the multi-task learning setting \cite{liu2021role}, ignoring the grounded relationships among the other causal variables of MHCH, like the user state and human cost.

To address above issues and improve the performance of MHCH, we propose a general Causal-Enhance Module (CEM), which can be plugged into existing MHCH networks \cite{liu2021time, liu2021role}, to incorporate the considerations of other potential causal variables of MHCH.
Specifically, we first analyzes MHCH task based on causal graph by mining all potential causal variables and deduce that user states and labor cost are the other two causal variables that should be considered for high-quality customer service. 
Then, to incorporate the consideration of user state, we train a user state network mainly driven by local sentiments to maintain the changes of user state during the dialogue and adjust the handoff predictions by correcting the prediction bias according to the causal relationship between user states and handoff decisions. 
To consider the labor cost of customer service and reduce it as much as possible while maintaining the same service quality, we construct a counterfactual-based cost simulator to regress the cost of a dialogue as an auxiliary task which can make the MHCH backbone become cost-aware and minimize the labor cost as much as possible. 

The contributions of our CEM can be summarized as follows:
\begin{itemize}
\item We conduct causal analysis based on causal graph for MHCH and identify the other two causal variables: user state and human cost, which should be considered to build high-quality MHCH service. 
\item To consider the impact of user state, the user state is applied to correct the handoff prediction bias according to the causal relationship between user states and handoff decisions.
\item To minimize the labor cost of customer service while maintaining the same service quality, we construct a counterfactual-based cost simulator to regress the cost of a dialogue as an auxiliary task, which can make the MHCH backbone become cost-aware.
\end{itemize}

\section{Related Work}


\noindent\textbf{Machine-Human Chatting Handoff.} The research on MHCH is originated in 2018. Using the idea of reinforcement learning, \citet{2018Evorus} proposed a dialogue robot to choose an assistant. \citet{rajendran2019learning} utilize a reinforcement learning framework to maximize success rate and minimize human workload. \citet{liu2021time, liu2021role} regraded the MHCH as a classificagtion problem and focused on identifying which sentence should be transferred to the human service. 


\noindent\textbf{Causal inference and counterfactual learning.} For structural causal models \cite{Halpern2005Causes}, related studies \cite{2013Bayesian, 2014Proof, xia2021causal} utilize graph neural networks for directed acyclic graph structure learning. For Rubin causal models, \citet{2006Matched} and \citet{ bengio2019meta} use neural networks to approximate the propensity scores, matching weights, etc., which can satisfy the covariate balancing \cite{kallus2020deepmatch, kuang2017estimating}; 
The representation learning \cite{huang2020dianet,liang2020instance} can also be used to matched the covariate balance between the test group and the reference group \cite{shalit2017estimating, louizos2017causal, lu2020reconsidering}. 
Several studies \cite{yoon2018ganite,  yuan2019improving, liu2020general} uses counterfactual methods based on the generative models over the observed distributions to causal inference. 


\noindent\textbf{Multi-task learning in dialogue systems.} \citet{2020Learning} uses multi-task learning for auxiliary pre-training tasks of dialogue data. \citet{2020DCR} combines dialogue behavior recognition and sentiment classification. \citet{2021Multi} proposes a model which includes generation and classification tasks. 

\begin{figure*}[t]
  \centering
  \includegraphics[width=1.0\linewidth]{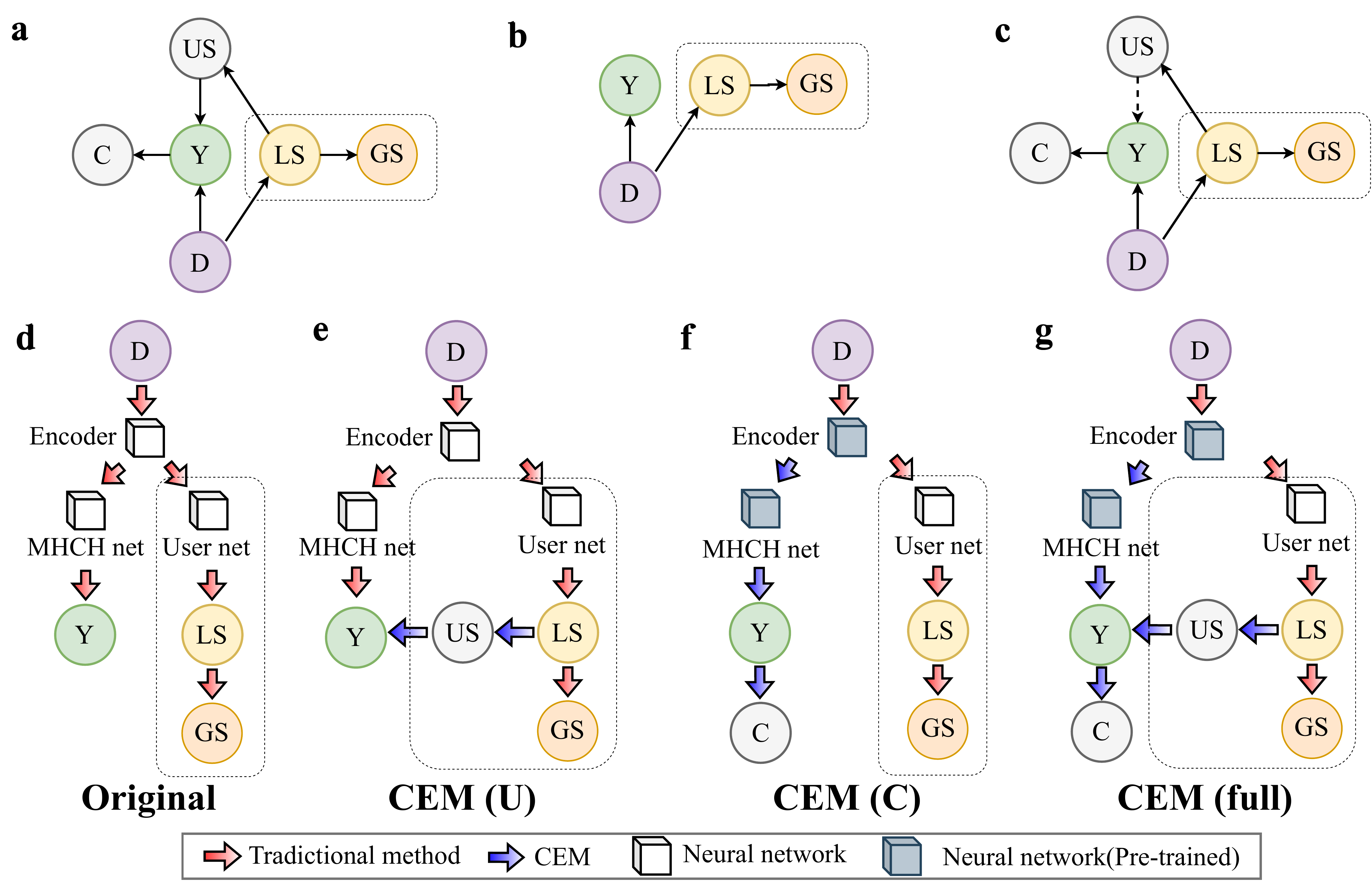}
  \caption{Causal graphs and model structures. D: Dialog, Y: prediction of MHCH, LS: local sentiment, GS: global satisfaction, US: user state, C: labor cost. The solid lines represent causality, the dashed line is adjustment, and the dotted line outlines the part that MHCH classification models doesn't have. \textbf{a, b, c} are the causal graphs of MHCH, traditional multi-task methods and CEM based on multi-task learning, respectively. \textbf{d} is the original model structure based on \textbf{b} \cite{song2019using, liu2021time, liu2021role}. \textbf{e}, \textbf{f} and \textbf{g} are the model structures enhanced by CEM(U), CEM(C) and CEM(full). }
\label{fig:causal_graph}
\end{figure*}

\section{Preliminary}
A given dialogue $D=[u_1,u_2,\ldots, u_L]$ contains $L$ utterances and have a label sequence $Y^h = [y_1^h, \ldots, y_L^h]$, where $y_t^h$ is the handoff label of $u_t$,$ 1 \leq t \leq L$.
The handoff labels $\mathrm{\Gamma}$ have two kinds of labels, i.e.,$\text{"normal"}$ and  $\text{"transferable"}$, where "normal" means that the utterance is no need to transfer, and "transferable" means that the utterance needs to be transferred to the manual service. 
The dialogue $D$ also have a global satisfaction label  $\{\text{"satisfactory"}, \text{"neutral"}, \text{"dissatisfied"}\}$. Then, the local sentiment of each utterance $u_t$ is measured by an open-source tool SnowNLP, which includes three labels $\{\text{"positive"}, \text{"neutral"}, \text{"negative"}\}$ .


\section{Methodology}
In this section, we analyse the impact of variables on MHCH from a fundamental view of causality. Then we present our CEM framework that eliminates the bad effect of ignored causal variables.


\subsection{Causal analysis of MHCH}

Causal graph is a directed acyclic graph where a node denotes a variable and an edge denotes a causal relation between two nodes \cite{pearl2009causality}. It is widely used to describe the process of data, which can guide the design of predictive models \cite{zhang2021causal}. Fig.\ref{fig:causal_graph}(a) shows the causal graph of MHCH. The rationality of this causal graph is explained as follows:
\begin{itemize}
\item D denote the dialogue $D=[u_1, \ldots, u_L]$.
\item Y = $[p_1,p_2,...,p_L]$ is the prediction of MHCH, where $p_t,1\leq t \leq L$ is the probability of that the handoff label of $u_t$ is $\text{"transferable"}$.
\item LS is the local sentiments of each utterance in a dialogue.
\item GS represents the user's subjective evaluation of the current dialogue.
\item US is a state for a given dialogue. Unlike GS, it is a variable that describes the objective state of the user. We can model US through local sentiments.
\item C is the labor cost caused by wrong prediction of MHCH.
\item Edge $D \rightarrow Y$: The MHCH can judge when to transfer to manual service according to the dialogue content. Therefore, the dialogue can affect the prediction of MHCH.  


\item Edge $Y \rightarrow C$: The labor cost depends on the prediction of MHCH. If we do not need to transfer to human service, there will not be labor cost.
\item Edges $D \rightarrow LS \rightarrow GS$: The dialogue quality of chatbot will affect users' sentiment, and then affect users' evaluation of the services.
\item Edge $LS \rightarrow US$: The user state can be modeled from local sentiments. 
\item Edge $US \rightarrow Y$: In Fig.\ref{fig:dialog}, the user state can affect MHCH to judge whether the service should be transferred to manual service.


\end{itemize}

However, instead of Fig.\ref{fig:causal_graph}(a), the existing common solutions, which are mainly based on multi-task methods, e.g., service satisfaction analysis (SSA) \cite{song2019using}, adopt the causal graph as Fig.\ref{fig:causal_graph}(b), which models the relationship of $D \rightarrow LS \rightarrow GS$. 
Specifically, they consider two neural networks for a multi-task of SSA and MHCH, i.e., train a user network (UN) for SSA and a MHCH network for MHCH as shown in Fig.\ref{fig:causal_graph}(d). Since there is an encoder network that share weights between the two tasks to integrate information, the local sentiment can assist MHCH network by sharing dialogue features. Although such modeling is simple and has good performance on MHCH tasks, it is established through a simplified causal graph without considering the factors of user and cost, so it can not completely show the overall picture as shown in Fig.\ref{fig:causal_graph}(a). Therefore, we design a new causal graph as seen in Fig.\ref{fig:causal_graph}(c) to consider further factors, e.g. user state and labor cost to bridge the MHCH network and UN. Based on the new causal graph, a novel CEM (full) model (Fig.\ref{fig:causal_graph}(g)) as well as its variants CEM(U) and CEM(C) will be introduced in the following sections.






\subsection{User State}




\begin{table*}[t]
  \centering
  \caption{Comparison of classification performance (\%) in Clothing and Makeup1. CEM-DAMI (C) refers to the model without the part in dashed box in Fig.\ref{fig:causal_graph}(f). We don't use full CEM to enhance DAMI for the reason that CEM cannot restore $US$ from pure MHCH models.}
  \resizebox*{\linewidth}{!}{
    \begin{tabular}{lcccccccccc}
    \toprule
    \multirow{2}[4]{*}{\textbf{Models}} & \multicolumn{5}{c}{\textbf{Clothing}} & \multicolumn{5}{c}{\textbf{Makeup1}} \\
\cmidrule{2-11} \multicolumn{1}{c}{} & \multicolumn{1}{c}{\textbf{F1}} & \multicolumn{1}{c}{\textbf{Mac.F1}} & \multicolumn{1}{c}{\textbf{GT-I}} & \multicolumn{1}{c}{\textbf{GT-II}} & \multicolumn{1}{c}{\textbf{GT-III}} & \multicolumn{1}{c}{\textbf{F1}} & \multicolumn{1}{c}{\textbf{Mac.F1}} & \multicolumn{1}{c}{\textbf{GT-I}} & \multicolumn{1}{c}{\textbf{GT-II}} & \multicolumn{1}{c}{\textbf{GT-III}} \\
    \midrule
    HRN \cite{lin2015hierarchical}   & 57.3  & 73.5  & 62.3  & 71.8  & 76.5  & 58.2  & 74.1  & 62.3  & 72.5  & 78.1 \\
    HAN \cite{yang2016hierarchical}   & 58.2  & 74.1  & 62.9  & 72.2  & 76.7  & 60.1  & 75.3  & 65.4  & 74.9  & 79.9 \\
    BERT \cite{devlin2018bert}  & 56.0  & 72.9  & 59.3  & 68.1  & 73.1  & 57.0  & 73.3  & 61.5  & 71.0  & 76.5 \\
    CRF-ASN \cite{chen2018dialogue} & 57.6  & 73.4  & 61.5  & 72.6  & 78.0  & 56.8  & 73.6  & 63.7  & 74.2  & 79.8 \\
    HBLSTM-CRF \cite{kumar2018dialogue} & 59.0  & 74.4  & 63.6  & 73.7  & 78.8  & 60.1  & 75.4  & 67.0  & 76.3  & 81.2 \\
    DialogueRNN \cite{majumder2019dialoguernn} & 59.0  & 74.3  & 63.1  & 73.8  & 79.0  & 61.3  & 76.1  & 66.3  & 76.0  & 81.2 \\
    CASA \cite{raheja2019dialogue}  & 59.7  & 74.7  & 64.8  & 74.9  & 79.7  & 60.4  & 75.7  & 67.8  & 77.0  & 81.8 \\
    LSTMLCA \cite{dai2020local} & 61.8  & 76.1  & 66.4  & 76.3  & 81.1  & 62.1  & 76.6  & 67.8  & 76.9  & 81.7 \\
    CESTa \cite{wang2020contextualized} & 60.5  & 75.2  & 64.0  & 74.6  & 79.6  & 60.2  & 75.2  & 65.2  & 75.9  & 81.5 \\
    DAMI \cite{liu2021time}  & 67.3  &  \textbf{79.7}&  \textbf{70.3}&  \textbf{79.1}&  \textbf{83.9}& 67.1  & 79.5  & 67.8  & 76.9  & 82.1 \\
    CEM-DAMI (C) &  \textbf{67.5}&  \textbf{79.7}& 69.7  & 77.6  & 81.5  &  \textbf{67.5}&  \textbf{79.9}&  \textbf{70.4}&  \textbf{78.1}& \textbf{82.2} \\
    \bottomrule
    \end{tabular}%
    }
  \label{tab:baselines1}%
\end{table*}%

As shown in Fig.\ref{fig:causal_graph}(e), we can use local sentiment, which is the output of UN, to restore user state. 
Since user state has a strong correlation with relative time \cite{ding2005time}, we can measure the user state of $u_t$ by Eq.(\ref{eq:us}) with the weighted sum of local sentiment.
\begin{equation}
    US_t = \sum^{L}_{t=1} \text{UN}(D) \times \beta_t,
    \label{eq:us}
\end{equation}
where $\text{UN}(D) \in R^{L \times 3}$ is the local sentiment from UN when given $D$ as input.
And the weight $\beta_t$ is
\begin{equation}
\resizebox{0.86\hsize}{!}{%
    $\beta_t=softmax([\frac{1}{L},\ldots,\frac{t-1}{L},\frac{t}{L}, 0, \ldots,0])$,
    }
\end{equation}


\textbf{Soft Adjustment. }In Fig.\ref{fig:causal_graph}(a), if we establish the causal relationship between user state and MHCH task directly, $D$ will become a confounder to $Y$ due to the intervention of user state. To solve this problem, a simple way is to ignore the causal relationship $US \rightarrow Y$. 

However, the utterance sometime can not affect directly  whether it is necessary to transfer to human service since the complexity of the language. And the user state restored by the local sentiment can help decision of MHCH. For example,in Fig.\ref{fig:dialog}, the information of the sentiment is more important than those of utterance. Therefore, we can use another strategy to use user state for adjustment the decision of MHCH network. In particular, we can mask the neutral local sentiment since the neutral sentiments are confusing and therefore not highly recognizable, which means that neutral sentiment's impact on MHCH \citet{song2019using} task is lower. 

Moreover, the dimensions of $US$ is three, which do not match the two-dimension output $Y$ of the MHCH network. While masking the neutral sentiments,
we can propose de-neutral soft adjustment shown in Fig.\ref{fig:causal_graph}(e), whose specific operation is as follows:
\begin{equation}
\resizebox{0.86\hsize}{!}{%
$y^h_D =  softmax(\text{Mask}_n(US_D) \odot \text{MHCH}(D)),$
}
\end{equation}
where $\text{Mask}_n$ is the masking operator for neutral sentiment.$\text{MHCH}$ is the model using to modeling the causal relationship of $D \rightarrow Y$. $y^h_D$ is the predicted result of $D$. $\odot$ represents a product operation at the element level, which makes the probability of "positive" times the probability of "normal" and makes the probability of "negative" times the probability of "transferable". This adjustment can modify the normal probability with positive sentiment and the transferable probability with negative sentiment.

\subsection{Cost Simulator}
Using $US$ can adjust MHCH models, eliminate the bias caused by not considering user factors, and obtain $Y$ that is closer to the ground truth. Based on this, we conduct counterfactual modeling of labor cost. Labor cost can be divided into two types, one is effective cost and the other is invalid cost. The effective cost refers to incur the cost for the utterances that must be transferred to human, and the invalid cost refers to incur the cost for the utterances that does not need to transfer to human, i.e., the cost caused by the wrong prediction. Note that we can not change the effective cost, and we can only reduce invalid cost. The network which models the causal relationship of $D \rightarrow Y \rightarrow C$ is named as cost simulator. We firstly define the cost simulator as follows:
\begin{equation}
\left\{
\begin{aligned}
&Y^h \sim P_{Y^h}(Y^h|D) \\
&C = F_c(Y^h, D),
\end{aligned}
\right.
\end{equation}
where $Y^h$ represents the ground truth of the MHCH task and $C$ denotes the labor cost of $D$. $P_{Y^h}$ is the probability calculation function for MHCH. $F_c$ represents the cost calculation function. $P_{Y^h}$ can be defined as follows:
\begin{equation}
P_{Y^h}(Y^h | D)=\prod_{t=1}^{L}{P(\hat{y}_t^h=y_t^h | u_t)},
\end{equation}
where $\hat{y}_t^h$ is the prediction of MHCH network for $u_t$, and $y_t^h$ is the ground truth of MHCH network for $u_t$. Then let $\zeta$ represent the upper limit of the cost of one utterance in human service. 
Since we only needs to estimate the relative cost, which means that it does not need to obtain a specific and accurate estimated value. Therefore, $\zeta$ is set to 1 by default. $F_c$ can be defined as follows:
\begin{equation}
F_c(Y^h,D)=\sum_{t=1}^L \zeta \cdot P(\hat{y}_t^h=1 | u_t).
\end{equation}
 In datasets, if the label corresponding to the data $u_t$ is "transferable", $P\left(y_t^h=1\middle| u_t\right)=1$, otherwise $P\left(y_t^h =1\middle| u_t\right)=0$. 
Next, we pretrain the cost simulator based on Eq.(\ref{eq:cost}), so that the predicted cost measured by the output of the MHCH network is close to the real labor cost. 
\begin{equation}
\mathcal{L}_{c_{pre}}=-MSE\left(\hat{C}-C\right),
\label{eq:cost}
\end{equation}
where $\hat{C}$ represents the cost predicted by the simulator, and $C$ represents the ground truth of cost. After supervised pre-training, the cost simulator can become cost-aware and can give a counterfactual cost. 

On the trained cost simulator, we begin to train the MHCH model, and calculate the counterfactual cost of each $D$ through $F_c$. Since we want to make labor cost as low as possible, the loss function $\mathcal{L}_c$ of the counterfactual cost simulator is defined as:
\begin{equation}
\begin{aligned}
\mathcal{L}_c&=-\sum_{i=1}^{|\mathrm{\Psi}|} \hat C \\
&=-\frac{1}{L}\sum_{i=1}^{|\mathrm{\Psi}|}\sum_{t=1}^{L}{\zeta \cdot P(y_{i,t}^h=1 | u_{i,t})}.
\end{aligned}
\end{equation}
where $|\Psi|$ is the dataset size.
Overall, the total loss $\mathcal{L}(\mathrm{\Theta})$ of CEM for the multi-task MHCH is 
\begin{equation}
\mathcal{L}(\mathrm{\Theta})=\mathcal{L}_h+\eta_s\cdot\mathcal{L}_s+\eta_c\cdot\mathcal{L}_c+\delta||\mathrm{\Theta}||_2^2,
\label{eq:totalloss}
\end{equation}
where $L_h$ and $L_s$ are the loss function of MHCH task and SSA task. $\eta_c$ and $\eta_s \in R^+$ are the weight parameters for multi tasks. the $\ell_2$ regularization $\delta||\mathrm{\Theta}||_2^2$ is used to mitigate model overfitting.

\section{Experiments}

\subsection{Dataset and Experimental Settings}
We evaluate our approach on four datasets including Clothing\cite{liu2021time}, Makeup1\cite{liu2021time}, Clothes\cite{liu2021role}, and Makeup2\cite{liu2021role}.  The statistics of the data are shown in Table \ref{tab:data}. To verify the effectiveness of CEM and fairly compare with the baselines on the same datasets, we evaluate the performance of our CEM on the classification model DAMI\cite{liu2021time} by using Clothing and Makeup1 while test the performance of CEM on the multi-task model RSSN\cite{liu2021role} with Clothes and Makeup2, for the reason that these models are state-of-the-art. Because DAMI only models the relationship of $D \rightarrow Y$, not $D \rightarrow LS \rightarrow GS$, it is not possible to use DAMI to get $US$, so we only add cost adjustment on DAMI named as CEM-DAMI (C).

\begin{table}[t]
  \centering
  \caption{Overall statistics of the datasets. }
  \resizebox*{\linewidth}{!}{
    \begin{tabular}{lcccc}
    \toprule
    \textbf{Statistics items} & \textbf{Clothing} & \textbf{Makeup1} & \textbf{Clothes} & \textbf{Makeup2} \\
    \midrule
    \# (Dialogues) & 3500  & 4000  & 10000 & 3540 \\
    \# (Dissatisfied dialogues) & -     & -     & 2302  & 1180 \\
    \# (Neutral dialogues) & -     & -     & 6399  & 1180 \\
    \# (Satisfactory dialogues) & -     & -     & 1299  & 1180 \\
    \# (Transferable utterances) & 6713  & 7446  & 16921 & 7668 \\
    \# (Normal utterances) & 28901 & 32488 & 237891 & 86778 \\
    Avg \# (Utterances per dialogues) & 10.18 & 9.98  & 25.48 & 26.68 \\
    \bottomrule
    \end{tabular}%
  }
  \label{tab:data}%
\end{table}%

\begin{table*}[htbp]
  \centering
  \caption{Comparison of classification performance (\%) in Clothes and Makeup2. CEM-RSSN (U) refers to the model in Fig.\ref{fig:causal_graph}(e), CEM-RSSN (C) refers to the model in Fig.\ref{fig:causal_graph}(f), and CEM-RSSN (full) refers to the model in Fig.\ref{fig:causal_graph}(g). }
    \resizebox*{\linewidth}{!}{
    \begin{tabular}{lcccccccccc}
    \toprule
    \multirow{2}[4]{*}{\textbf{Models}} & \multicolumn{5}{c}{\textbf{Clothes}} & \multicolumn{5}{c}{\textbf{Makeup2}} \\
\cmidrule{2-11}    \multicolumn{1}{c}{} & \multicolumn{1}{c}{\textbf{F1}} & \multicolumn{1}{c}{\textbf{Mac.F1}} & \multicolumn{1}{c}{\textbf{GT-I}} & \multicolumn{1}{c}{\textbf{GT-II}} & \multicolumn{1}{c}{\textbf{GT-III}} & \multicolumn{1}{c}{\textbf{F1}} & \multicolumn{1}{c}{\textbf{Mac.F1}} & \multicolumn{1}{c}{\textbf{GT-I}} & \multicolumn{1}{c}{\textbf{GT-II}} & \multicolumn{1}{c}{\textbf{GT-III}} \\
    \midrule
    HAN \cite{yang2016hierarchical}  & 59.8  & 78.7  & 71.7  & 73.1  & 74    & 54.3  & 75.4  & 68.5  & 70.1  & 71.3 \\
    BERT+LSTM \cite{devlin2018bert} & 60.4  & 78.9  & 73.4  & 74.9  & 75.9  & 42.2  & 84.2  & 72.9  & 66.4  & 77.6 \\
    HEC \cite{kumar2018dialogue}   & 59.8  & 78.7  & 71.2  & 72.3  & 73    & 57.1  & 76.8  & 68    & 69.5  & 70.5 \\
    DialogueRNN \cite{majumder2019dialoguernn} & 60.8  & 79.2  & 73.1  & 74.6  & 75.6  & 58.3  & 77.4  & 68.8  & 70.5  & 71.6 \\
    CASA \cite{raheja2019dialogue} & 62    & 79.8  & 73.6  & 75    & 75.9  & 58.4  & 77.5  & 70.6  & 72.7  & 73.9 \\
    LSTMLCA \cite{dai2020local} & 62.6  & 80.1  & 72.4  & 73.9  & 74.8  & 57.4  & 77    & 70.2  & 71.7  & 72.6 \\
    CESTa \cite{wang2020contextualized} & 60.6  & 79.1  & 73.4  & 74.8  & 75.6  & 59.3  & 78    & 69.6  & 71.2  & 72.2 \\
    DAMI \cite{liu2021time}  & 66.7  & 82.2  & 74.2  & 75.9  & 77.1  & 61.1  & 79    & 73.3  & 74.4  & 75.2 \\
    \midrule
    MT-ES \cite{ma2018detect} & 61.7  & 79.7  & 74.6  & 75.9  & 76.8  & 57.1  & 76.9  & 69.9  & 71.7  & 72.8 \\
    JointBiLSTM \cite{bodigutla2020joint} & 62    & 79.9  & 75    & 76.1  & 76.9  & 59.3  & 78    & 70.1  & 72    & 73.1 \\
    DCR-Net \cite{2020DCR} & 62.1  & 79.9  & 71.4  & 72.8  & 73.7  & 58.8  & 77.7  & 70    & 72.1  & 73.4 \\
    RSSN \cite{liu2021role}   & 67.1  & 82.5  & 75.8  & 77.1  & 78    & 63.5  & 80.2  & 74.9  & 76.5  & 77.7 \\
    \midrule
    CEM-RSSN (U) & 66.6  & 82.4  & 79.8  & 80.5  & 81    & 67.6  & 82.8  & 80.2  & 81.1  & 81.8 \\
    CEM-RSSN (C) & 66.1  & 82.1  & 78.6  & 79.6  & 80.2  & 65.2  & 81.6  & 77.3  & 78.2  & 78.9 \\
    CEM-RSSN (full) & 67.9  & 82.9  & 79.6  & 80.7  & 81.4  & 64.8  & 80.9  & 79.4  & 81.1  & 82.3 \\
    \bottomrule
    \end{tabular}%
    }
  \label{tab:baselines2}%
\end{table*}%




\subsection{Evaluation Metrics}
Following prior works \citet{liu2021time,liu2021role}, we adopt F1, Macro F1 (Mac.F1) and GT-T (Golden Transfer within Tolerance) as accuracy metrics for evaluating the MHCH task. GT-T takes into account the tolerance property of the MHCH task through a tolerance range T, which allows for "biased" predictions within it. The T can be ranged from 1 to 3 corresponding to GT-I, GT-II, and GT-III.

Furthermore, to verify that CEM can effectively control the cost, we compare the labor cost of different models. It is obvious that the higher the accuracy, the lower the labor cost. To eliminate the impact of model accuracy, we only compare the invalid cost which is more meaningful than the full cost. Therefore, we compute the invalid cost as follows:
\begin{equation}
\resizebox{0.81\hsize}{!}{%
$\text{IC} = \frac{\sum_{i=1}^{|\mathrm{\Psi}|}\sum_{t=1}^{L}(\hat{y}_{i,t}^h != y_{i,t}^h,\ \hat{y}_{i, t}^h = \text{"transferable"})}{\sum_{i=1}^{|\mathrm{\Psi}|}\sum_{t=1}^{L}(\hat{y}_{i,t}^h != y_{i,t}^h)}$
}
\label{eq:cost}
\end{equation}
where IC is the abbreviation of invalid cost.




\subsection{Implementation Details}
We use TensorFlow\footnote{https://www.tensorflow.org/} to implement our method with one RTX2080 GPU card. Back-propagation is used to compute gradients and the Adam optimizer \cite{kingma2014adam} is used for parameter updates. The dimension of word embedding is set as 200. The total vocabulary size of datasets is 48.5K. Other trainable model parameters are initialized by sampling values from initializer. Hyper-parameters of CEM and baselines are tuned on the validation set. $\eta_s$ is set as 0.3. The sizes of model units are based on the baselines setting and remain the same in the comparison experiments. The $L_2$ regularization weight is $10^{-4}$ in DAMI and $3 \times 10^{-5}$ in RSSN. The batch size is set as 32. The number of epochs is set as 30 in DAMI and CEM-DAMI(C), and set as 80 in RSSN, CEM-RSSN(C), CEM-RSSN(U) and CEM-RSSN. Finally, we train the models with a learning rate of $7.5 \times 10^{-3}$ in DAMI and $1.5 \times 10^{-3}$ in RSSN. Following the data processing setting in \citet{liu2021time}, the datasets are divided into training set, validation set, and test set with an ratio of 8:1:1.

\subsection{Results on Clothing and Makeup1}

The experimental results of models on Clothing and Makeup1 are shown in Table \ref{tab:baselines1}. In Clothing, CEM-DAMI outperforms most baselines, and is slightly weaker than DAMI on GT-T metrics. In Makeup1, CEM-DAMI is the best performing model. This experimental result shows that incorporating cost into models does not reduce model accuracy. 

$IC$ of DAMI with the adjustment of CEM is significantly reduced in both Clothing and Makeup1 as shown in Table \ref{tab:cost}. We can conclude based on Table \ref{tab:baselines1} and Table \ref{tab:cost} that CEM-DAMI(C) can achieve competitive performance with lower labor cost, which means that our cost simulator can reduce labor cost while maintaining the model performance. 

\begin{table}[htbp]
  \centering
  \caption{Comparison of IC (\%) in Clothing and Makeup1.}
    \begin{tabular}{lcc}
    \toprule
    \textbf{Models} & \multicolumn{1}{c}{\textbf{Clothing}} & \multicolumn{1}{c}{\textbf{Makeup1}} \\
    \midrule
    DAMI  & 59.0  & 56.6 \\
    CEM-DAMI (C) & 53.4  & 54.2 \\
    \bottomrule
    \end{tabular}%
  \label{tab:cost}%
\end{table}%



\subsection{Results on Clothes and Makeup2}

\begin{figure*}
  \centering
  \includegraphics[width=0.9\linewidth]{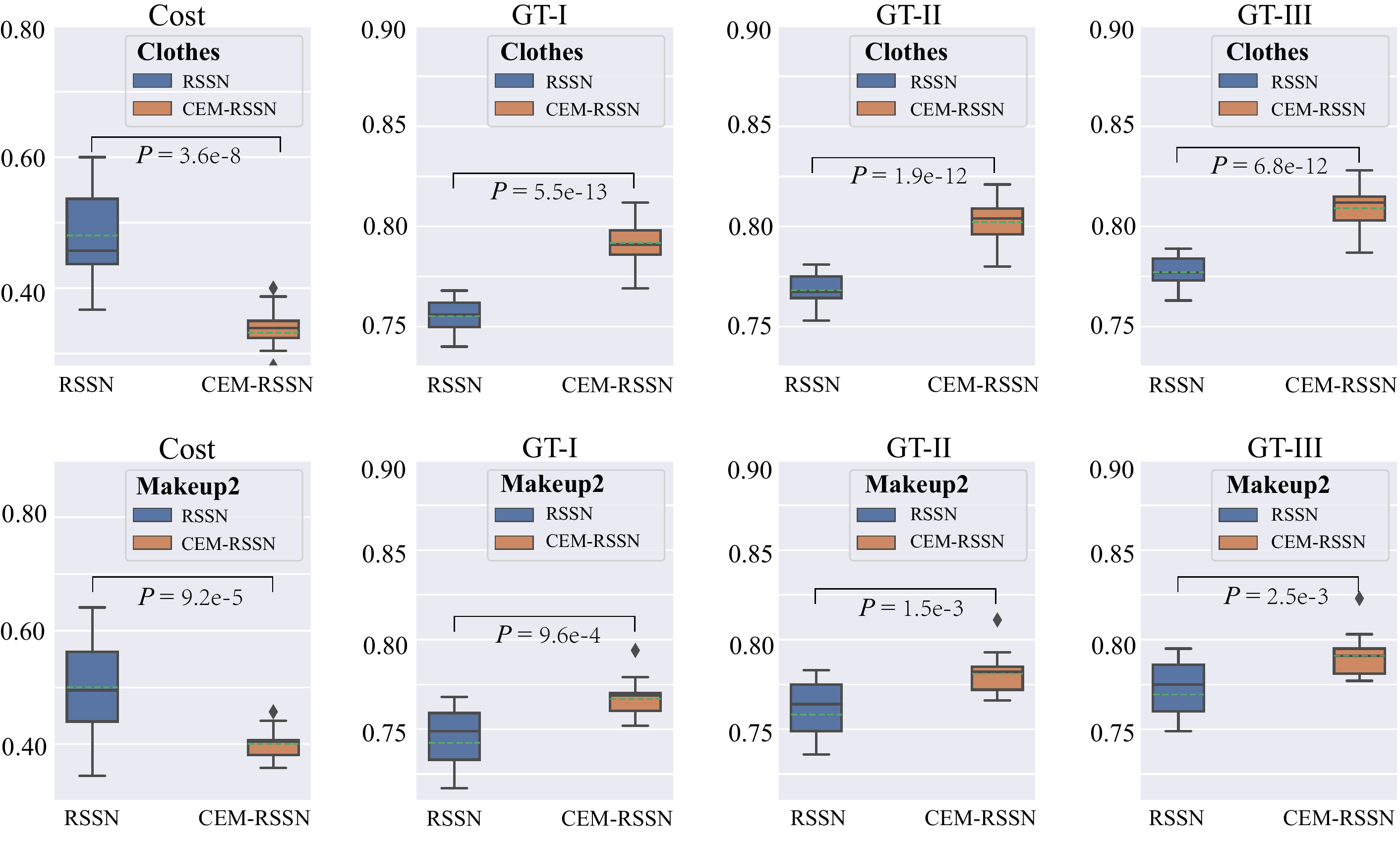}
  \caption{Comparison of labor cost (\%) and GT-T in Clothes and Makeup2. Cost is IC defined in Eq.(\ref{eq:cost}). We obtain multiple sets of results of RSSN and CEM-RSSN through multiple experiments, draw the box plots of Cost and GT-T, and perform two sided t-test to verify the whether the significant difference between RSSN and CEM-RSSN over four metrics. }
\label{fig:cost}
\end{figure*}

The experimental results of the methods on Clothes and Makeup2 are shown in Table \ref{tab:baselines2}. The cost results are shown in Figure \ref{fig:cost}.  From Table \ref{tab:baselines2}, we can observe that our CEM can effectively improve the performance of RSSN, especially on GT-I, GT-II, and GT-III.
We investigate the effects of user state and cost simulator through ablation experiments. According to the experimental results shown in Table \ref{tab:baselines2}, CEM-RSSN (U) can effectively improve model performance by modeling and tracking user state which is highly-correlated with user's tolerances for invalid responses. Besides, similar with the results on Clothing and Makeup1, CEM-RSSN (C) still can achieve competitive performance even the cost simulator is not designed to improve the accuracy of the model.   

Finally, We also compare the cost of RSSN and CEM-RSSN through multiple experiments, and perform two sided t-test to verify the whether the significant difference between RSSN and CEM-RSSN over metrics. The results shown in Fig.\ref{fig:cost} mean that CEM can significantly reduce labor cost while improving model performance.




\subsection{Parameter Sensitivity}

\begin{figure}
  \centering
  \includegraphics[width=0.99\linewidth]{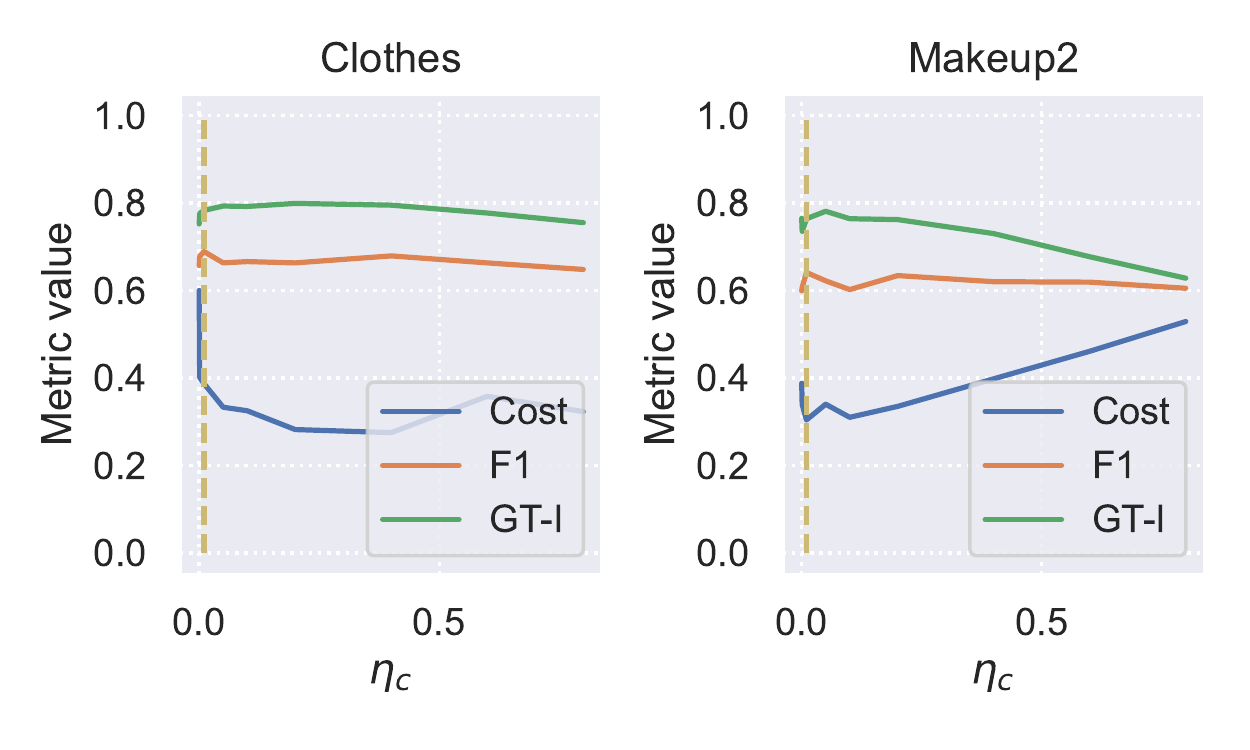}
  \caption{The impact of the cost simulator over different $\eta_c\in[0,1]$. This experiment is about CEM-RSSN on Clothes and Makeup2 dataset. Cost is IC which is defined in Eq.(\ref{eq:cost}). F1 and GT-I is the metrics about the MHCH accuracy. }
\label{fig:loss_weight}
\end{figure}

We compare different weight $\eta_c$ in Eq.\ref{eq:totalloss} to explore the impact of the cost simulator on MHCH models. 
 We take the value of $\eta_c$ from 0 to 1, and the experimental results about CEM-RSSN on Clothes and Makeup2 are shown in Fig.\ref{fig:loss_weight}. 
It can be seen that the different $\eta_c$ has little effect on the green (GT-I) and orange lines (F1) , which indicates the stability of CEM in cost control. 

Comparing the metric values under different weight $\eta_c$ on both two datasets, it can be found that when the weight exceeds a certain threshold, the larger the weight, the higher the labor cost and the lower the accuracy, indicating the labor cost is associated with the accuracy of models. Besides, it also shows that the trade-off between the accuracy and labor cost can be achieved by adjusting the loss weight, so as to obtain a model with low cost and high accuracy. 
Meanwhile, when the $\eta_c$ is lower than a certain threshold, the effect of the cost simulator can be ignored, which will affect the performance of CEM. Finally, we chose 0.01 as the value of $\eta_c$ to make a better trade-off in our experiments.


\section{Conclusions}

In this paper, we propose a novel CEM module for the MHCH task where we use causal inference to enhance the models of the MHCH task, and takes into account the labor cost. And the empirical results on four datastes and two types of models indicate that CEM improves models accuracy consistently and effectively saves invalid labor cost. 
Since there are minor modifications to the model architecture and loss function on the existing MHCH method and achieve the significant improvement, CEM can be easily plugged into the different MHCH methods. 

\section{Future works}
Based on CEM, we can consider the following future work:

\noindent (1) In fact, when CEM is used to enhance the existing MHCH method, there is no additional parameters. From the perspective of inference speed and model deployment, this is the advantage. However, the consideration of user state and labor cost in CEM is mainly based on intuition to construct explicit transformations, which can not ensure the good enough performance. Therefore, we can consider adding neural networks to CEM in the future.

\noindent (2) For MHCH task, the conventional neural networks are generally used, such as fully connected neural networks, LSTM, BiLSTM etc. This makes the structure of the model lack of careful consideration, which has the potential to greatly improve the performance of MHCH. Specifically, we should consider a lot of fine-tuning methods for the neural network to ensure their performance, including the use of structure search techniques \cite{DBLP:conf/icann/HeHLLY21,liu2018progressive,huang2020efficient}, elaborate modules \cite{hu2018squeeze,huang2020dianet}, specific parameters \cite{liang2020instance}, etc.

\noindent (3) Compared with other artificial intelligence fields, such as image segmentation and image classification, the data volume shown in Table.\ref{tab:data} is not very large. However, the quality and quantity of data have a huge impact on their training, so the data driven cost simulator may have bias in its estimation of labor cost. Therefore, we can consider data augment methods\cite{DBLP:journals/corr/abs-1805-09501,lin2021continuous} to effectively improve model training.

\section*{Limitations}

Although we have fully demonstrated the effectiveness of CEM experimentally, we ignore the analysis for CEM from the mathematical point of view of causal inference. This makes it impossible for us to guarantee that CEM can be used in more complex and sophisticated MHCH methods in the future, or other applications in more extensive fields. Moreover, since the cost simulator is trained by neural network, we can not ensure whether the cost given by the simulator can not have a well enough performance to reflect the truth labor cost.



\bibliography{anthology,custom}

\begin{thebibliography}{50}
\expandafter\ifx\csname natexlab\endcsname\relax\def\natexlab#1{#1}\fi

\bibitem[{Bengio et~al.(2019)Bengio, Deleu, Rahaman, Ke, Lachapelle, Bilaniuk,
  Goyal, and Pal}]{bengio2019meta}
Yoshua Bengio, Tristan Deleu, Nasim Rahaman, Rosemary Ke, S{\'e}bastien
  Lachapelle, Olexa Bilaniuk, Anirudh Goyal, and Christopher Pal. 2019.
\newblock A meta-transfer objective for learning to disentangle causal
  mechanisms.
\newblock \emph{arXiv preprint arXiv:1901.10912}.

\bibitem[{Bodigutla et~al.(2020)Bodigutla, Tiwari, Vargas, Polymenakos, and
  Matsoukas}]{bodigutla2020joint}
Praveen~Kumar Bodigutla, Aditya Tiwari, Josep~Valls Vargas, Lazaros
  Polymenakos, and Spyros Matsoukas. 2020.
\newblock Joint turn and dialogue level user satisfaction estimation on
  multi-domain conversations.
\newblock \emph{arXiv preprint arXiv:2010.02495}.

\bibitem[{Chen et~al.(2018)Chen, Yang, Zhao, Cai, and He}]{chen2018dialogue}
Zheqian Chen, Rongqin Yang, Zhou Zhao, Deng Cai, and Xiaofei He. 2018.
\newblock Dialogue act recognition via crf-attentive structured network.
\newblock In \emph{The 41st international acm sigir conference on research \&
  development in information retrieval}, pages 225--234.

\bibitem[{Claassen et~al.(2014)Claassen, Mooij, and Heskes}]{2014Proof}
T.~Claassen, J.~M. Mooij, and T.~Heskes. 2014.
\newblock Proof supplement - learning sparse causal models is not np-hard
  (uai2013).
\newblock \emph{Statistics}.

\bibitem[{Cubuk et~al.(2018)Cubuk, Zoph, Mane, Vasudevan, and
  Le}]{DBLP:journals/corr/abs-1805-09501}
Ekin~Dogus Cubuk, Barret Zoph, Dandelion Mane, Vijay Vasudevan, and Quoc~V. Le.
  2018.
\newblock \href {http://arxiv.org/abs/1805.09501} {Autoaugment: Learning
  augmentation policies from data}.
\newblock \emph{CoRR}, abs/1805.09501.

\bibitem[{Dai et~al.(2020)Dai, Fu, Zhu, Cui, Qi et~al.}]{dai2020local}
Zhigang Dai, Jinhua Fu, Qile Zhu, Hengbin Cui, Yuan Qi, et~al. 2020.
\newblock Local contextual attention with hierarchical structure for dialogue
  act recognition.
\newblock \emph{arXiv preprint arXiv:2003.06044}.

\bibitem[{Devlin et~al.(2018)Devlin, Chang, Lee, and
  Toutanova}]{devlin2018bert}
Jacob Devlin, Ming-Wei Chang, Kenton Lee, and Kristina Toutanova. 2018.
\newblock Bert: Pre-training of deep bidirectional transformers for language
  understanding.
\newblock \emph{arXiv preprint arXiv:1810.04805}.

\bibitem[{Ding and Li(2005)}]{ding2005time}
Yi~Ding and Xue Li. 2005.
\newblock Time weight collaborative filtering.
\newblock In \emph{Proceedings of the 14th ACM international conference on
  Information and knowledge management}, pages 485--492.

\bibitem[{Halpern et~al.(2005)Halpern, Joseph, Y., Pearl, and
  Judea}]{Halpern2005Causes}
Halpern, Joseph, Y., Pearl, and Judea. 2005.
\newblock Causes and explanations: A structural-model approach. part i: Causes.
\newblock \emph{British Journal for the Philosophy of Science}.

\bibitem[{He et~al.(2016)He, Zhang, Ren, and Sun}]{He_2016_CVPR}
Kaiming He, Xiangyu Zhang, Shaoqing Ren, and Jian Sun. 2016.
\newblock Deep residual learning for image recognition.
\newblock In \emph{Proceedings of the IEEE Conference on Computer Vision and
  Pattern Recognition (CVPR)}.

\bibitem[{He et~al.(2021)He, Huang, Liang, Liang, and
  Yang}]{DBLP:conf/icann/HeHLLY21}
Wei He, Zhongzhan Huang, Mingfu Liang, Senwei Liang, and Haizhao Yang. 2021.
\newblock Blending pruning criteria for convolutional neural networks.
\newblock In \emph{Artificial Neural Networks and Machine Learning - {ICANN}
  2021 - 30th International Conference on Artificial Neural Networks,
  Bratislava, Slovakia, September 14-17, 2021, Proceedings, Part {IV}}, volume
  12894 of \emph{Lecture Notes in Computer Science}, pages 3--15. Springer.

\bibitem[{Heskes(2013)}]{2013Bayesian}
T.~Heskes. 2013.
\newblock Bayesian probabilities for constraint-based causal discovery.

\bibitem[{Hu et~al.(2018)Hu, Shen, and Sun}]{hu2018squeeze}
Jie Hu, Li~Shen, and Gang Sun. 2018.
\newblock Squeeze-and-excitation networks.
\newblock In \emph{Proceedings of the IEEE conference on computer vision and
  pattern recognition}, pages 7132--7141.

\bibitem[{Huang et~al.(2018)Huang, Chang, and Bigham}]{2018Evorus}
Ting Hao~'Kenneth' Huang, Joseph~Chee Chang, and Jeffrey~P. Bigham. 2018.
\newblock Evorus: A crowd-powered conversational assistant built to automate
  itself over time.
\newblock \emph{ACM}.

\bibitem[{Huang et~al.(2020{\natexlab{a}})Huang, Liang, Liang, He, and
  Yang}]{huang2020efficient}
Zhongzhan Huang, Senwei Liang, Mingfu Liang, Wei He, and Haizhao Yang.
  2020{\natexlab{a}}.
\newblock Efficient attention network: Accelerate attention by searching where
  to plug.
\newblock \emph{arXiv preprint arXiv:2011.14058}.

\bibitem[{Huang et~al.(2020{\natexlab{b}})Huang, Liang, Liang, and
  Yang}]{huang2020dianet}
Zhongzhan Huang, Senwei Liang, Mingfu Liang, and Haizhao Yang.
  2020{\natexlab{b}}.
\newblock Dianet: Dense-and-implicit attention network.
\newblock In \emph{Proceedings of the AAAI Conference on Artificial
  Intelligence}, volume~34, pages 4206--4214.

\bibitem[{Ide and Kawahara(2021)}]{2021Multi}
T.~Ide and D.~Kawahara. 2021.
\newblock Multi-task learning of generation and classification for
  emotion-aware dialogue response generation.

\bibitem[{Kallus(2020)}]{kallus2020deepmatch}
Nathan Kallus. 2020.
\newblock Deepmatch: Balancing deep covariate representations for causal
  inference using adversarial training.
\newblock In \emph{International Conference on Machine Learning}, pages
  5067--5077. PMLR.

\bibitem[{Kingma and Ba(2014)}]{kingma2014adam}
Diederik~P Kingma and Jimmy Ba. 2014.
\newblock Adam: A method for stochastic optimization.
\newblock \emph{arXiv preprint arXiv:1412.6980}.

\bibitem[{Kuang et~al.(2017)Kuang, Cui, Li, Jiang, and
  Yang}]{kuang2017estimating}
Kun Kuang, Peng Cui, Bo~Li, Meng Jiang, and Shiqiang Yang. 2017.
\newblock Estimating treatment effect in the wild via differentiated confounder
  balancing.
\newblock In \emph{Proceedings of the 23rd ACM SIGKDD international conference
  on knowledge discovery and data mining}, pages 265--274.

\bibitem[{Kumar et~al.(2018)Kumar, Agarwal, Dasgupta, and
  Joshi}]{kumar2018dialogue}
Harshit Kumar, Arvind Agarwal, Riddhiman Dasgupta, and Sachindra Joshi. 2018.
\newblock Dialogue act sequence labeling using hierarchical encoder with crf.
\newblock In \emph{Proceedings of the aaai conference on artificial
  intelligence}, volume~32.

\bibitem[{Liang et~al.(2020)Liang, Huang, Liang, and Yang}]{liang2020instance}
Senwei Liang, Zhongzhan Huang, Mingfu Liang, and Haizhao Yang. 2020.
\newblock Instance enhancement batch normalization: An adaptive regulator of
  batch noise.
\newblock In \emph{Proceedings of the AAAI Conference on Artificial
  Intelligence}, volume~34, pages 4819--4827.

\bibitem[{Liang et~al.(2022)Liang, Huang, and Zhang}]{liang2022stiffnessaware}
SENWEI Liang, Zhongzhan Huang, and Hong Zhang. 2022.
\newblock Stiffness-aware neural network for learning hamiltonian systems.
\newblock In \emph{International Conference on Learning Representations}.

\bibitem[{Lin et~al.(2021)Lin, Huang, Wang, Liang, Chen, and
  Lin}]{lin2021continuous}
Junfan Lin, Zhongzhan Huang, Keze Wang, Xiaodan Liang, Weiwei Chen, and Liang
  Lin. 2021.
\newblock Continuous transition: Improving sample efficiency for continuous
  control problems via mixup.
\newblock In \emph{2021 IEEE International Conference on Robotics and
  Automation (ICRA)}, pages 9490--9497. IEEE.

\bibitem[{Lin et~al.(2015)Lin, Liu, Yang, Li, Zhou, and
  Li}]{lin2015hierarchical}
Rui Lin, Shujie Liu, Muyun Yang, Mu~Li, Ming Zhou, and Sheng Li. 2015.
\newblock Hierarchical recurrent neural network for document modeling.
\newblock In \emph{Proceedings of the 2015 Conference on Empirical Methods in
  Natural Language Processing}, pages 899--907.

\bibitem[{Liu et~al.(2018)Liu, Zoph, Neumann, Shlens, Hua, Li, Fei-Fei, Yuille,
  Huang, and Murphy}]{liu2018progressive}
Chenxi Liu, Barret Zoph, Maxim Neumann, Jonathon Shlens, Wei Hua, Li-Jia Li,
  Li~Fei-Fei, Alan Yuille, Jonathan Huang, and Kevin Murphy. 2018.
\newblock Progressive neural architecture search.
\newblock In \emph{Proceedings of the European conference on computer vision
  (ECCV)}, pages 19--34.

\bibitem[{Liu et~al.(2020)Liu, Cheng, Dong, He, Pan, and Ming}]{liu2020general}
Dugang Liu, Pengxiang Cheng, Zhenhua Dong, Xiuqiang He, Weike Pan, and Zhong
  Ming. 2020.
\newblock A general knowledge distillation framework for counterfactual
  recommendation via uniform data.
\newblock In \emph{Proceedings of the 43rd International ACM SIGIR Conference
  on Research and Development in Information Retrieval}, pages 831--840.

\bibitem[{Liu et~al.(2021{\natexlab{a}})Liu, Gao, Kang, Jiang, He, Sun, Liu,
  and Lu}]{liu2021time}
Jiawei Liu, Zhe Gao, Yangyang Kang, Zhuoren Jiang, Guoxiu He, Changlong Sun,
  Xiaozhong Liu, and Wei Lu. 2021{\natexlab{a}}.
\newblock Time to transfer: Predicting and evaluating machine-human chatting
  handoff.
\newblock In \emph{Proc. of AAAI}, pages 5841--5849.

\bibitem[{Liu et~al.(2021{\natexlab{b}})Liu, Song, Kang, He, Jiang, Sun, Lu,
  and Liu}]{liu2021role}
Jiawei Liu, Kaisong Song, Yangyang Kang, Guoxiu He, Zhuoren Jiang, Changlong
  Sun, Wei Lu, and Xiaozhong Liu. 2021{\natexlab{b}}.
\newblock A role-selected sharing network for joint machine-human chatting
  handoff and service satisfaction analysis.
\newblock \emph{arXiv preprint arXiv:2109.08412}.

\bibitem[{Louizos et~al.(2017)Louizos, Shalit, Mooij, Sontag, Zemel, and
  Welling}]{louizos2017causal}
Christos Louizos, Uri Shalit, Joris~M Mooij, David Sontag, Richard Zemel, and
  Max Welling. 2017.
\newblock Causal effect inference with deep latent-variable models.
\newblock \emph{Advances in neural information processing systems}, 30.

\bibitem[{Lu et~al.(2020)Lu, Tao, Chen, Li, Guo, and
  Carin}]{lu2020reconsidering}
Danni Lu, Chenyang Tao, Junya Chen, Fan Li, Feng Guo, and Lawrence Carin. 2020.
\newblock Reconsidering generative objectives for counterfactual reasoning.
\newblock \emph{Advances in Neural Information Processing Systems},
  33:21539--21553.

\bibitem[{Ma et~al.(2018)Ma, Gao, and Wong}]{ma2018detect}
Jing Ma, Wei Gao, and Kam-Fai Wong. 2018.
\newblock Detect rumor and stance jointly by neural multi-task learning.
\newblock In \emph{Companion proceedings of the the web conference 2018}, pages
  585--593.

\bibitem[{Majumder et~al.(2019)Majumder, Poria, Hazarika, Mihalcea, Gelbukh,
  and Cambria}]{majumder2019dialoguernn}
Navonil Majumder, Soujanya Poria, Devamanyu Hazarika, Rada Mihalcea, Alexander
  Gelbukh, and Erik Cambria. 2019.
\newblock Dialoguernn: An attentive rnn for emotion detection in conversations.
\newblock In \emph{Proceedings of the AAAI Conference on Artificial
  Intelligence}, volume~33, pages 6818--6825.

\bibitem[{Pearl(2009)}]{pearl2009causality}
Judea Pearl. 2009.
\newblock \emph{Causality}.
\newblock Cambridge university press.

\bibitem[{Qin et~al.(2020)Qin, Che, Li, Ni, and Liu}]{2020DCR}
L.~Qin, W.~Che, Y.~Li, M.~Ni, and T.~Liu. 2020.
\newblock Dcr-net: A deep co-interactive relation network for joint dialog act
  recognition and sentiment classification.
\newblock \emph{Proceedings of the AAAI Conference on Artificial Intelligence},
  34(5):8665--8672.

\bibitem[{Radziwill and Benton(2017)}]{2017Evaluating}
Nicole Radziwill and Morgan Benton. 2017.
\newblock Evaluating quality of chatbots and intelligent conversational agents.
\newblock \emph{Software Quality Professional}, 19(3):25--35.

\bibitem[{Raheja and Tetreault(2019)}]{raheja2019dialogue}
Vipul Raheja and Joel Tetreault. 2019.
\newblock Dialogue act classification with context-aware self-attention.
\newblock \emph{arXiv preprint arXiv:1904.02594}.

\bibitem[{Rajendran et~al.(2019)Rajendran, Ganhotra, and
  Polymenakos}]{rajendran2019learning}
Janarthanan Rajendran, Jatin Ganhotra, and Lazaros~C Polymenakos. 2019.
\newblock Learning end-to-end goal-oriented dialog with maximal user task
  success and minimal human agent use.
\newblock \emph{Transactions of the Association for Computational Linguistics},
  7:375--386.

\bibitem[{Ren et~al.(2015)Ren, He, Girshick, and Sun}]{ren2015faster}
Shaoqing Ren, Kaiming He, Ross Girshick, and Jian Sun. 2015.
\newblock Faster r-cnn: Towards real-time object detection with region proposal
  networks.
\newblock \emph{Advances in neural information processing systems}, 28.

\bibitem[{Rubin(2006)}]{2006Matched}
D.~B. Rubin. 2006.
\newblock \emph{Matched Sampling for Causal Effects}.
\newblock Matched sampling for causal effects /.

\bibitem[{Shalit et~al.(2017)Shalit, Johansson, and
  Sontag}]{shalit2017estimating}
Uri Shalit, Fredrik~D Johansson, and David Sontag. 2017.
\newblock Estimating individual treatment effect: generalization bounds and
  algorithms.
\newblock In \emph{International Conference on Machine Learning}, pages
  3076--3085. PMLR.

\bibitem[{Song et~al.(2019)Song, Bing, Gao, Lin, Zhao, Wang, Sun, Liu, and
  Zhang}]{song2019using}
Kaisong Song, Lidong Bing, Wei Gao, Jun Lin, Lujun Zhao, Jiancheng Wang,
  Changlong Sun, Xiaozhong Liu, and Qiong Zhang. 2019.
\newblock Using customer service dialogues for satisfaction analysis with
  context-assisted multiple instance learning.

\bibitem[{Wang et~al.(2020)Wang, Zhang, Ma, Wang, and
  Xiao}]{wang2020contextualized}
Yan Wang, Jiayu Zhang, Jun Ma, Shaojun Wang, and Jing Xiao. 2020.
\newblock Contextualized emotion recognition in conversation as sequence
  tagging.
\newblock In \emph{Proceedings of the 21th Annual Meeting of the Special
  Interest Group on Discourse and Dialogue}, pages 186--195.

\bibitem[{Xia et~al.(2021)Xia, Lee, Bengio, and Bareinboim}]{xia2021causal}
Kevin Xia, Kai-Zhan Lee, Yoshua Bengio, and Elias Bareinboim. 2021.
\newblock The causal-neural connection: Expressiveness, learnability, and
  inference.
\newblock \emph{Advances in Neural Information Processing Systems}, 34.

\bibitem[{Xu et~al.(2020)Xu, Tao, Jiang, Zhao, Zhao, and Yan}]{2020Learning}
R.~Xu, C.~Tao, D~Jiang, X.~Zhao, D~Zhao, and R.~Yan. 2020.
\newblock Learning an effective context-response matching model with
  self-supervised tasks for retrieval-based dialogues.

\bibitem[{Xu et~al.(2019)Xu, Zhang, Luo, Xiao, and Ma}]{xu2019frequency}
Zhi-Qin~John Xu, Yaoyu Zhang, Tao Luo, Yanyang Xiao, and Zheng Ma. 2019.
\newblock Frequency principle: Fourier analysis sheds light on deep neural
  networks.
\newblock \emph{arXiv preprint arXiv:1901.06523}.

\bibitem[{Yang et~al.(2016)Yang, Yang, Dyer, He, Smola, and
  Hovy}]{yang2016hierarchical}
Zichao Yang, Diyi Yang, Chris Dyer, Xiaodong He, Alex Smola, and Eduard Hovy.
  2016.
\newblock Hierarchical attention networks for document classification.
\newblock In \emph{Proceedings of the 2016 conference of the North American
  chapter of the association for computational linguistics: human language
  technologies}, pages 1480--1489.

\bibitem[{Yoon et~al.(2018)Yoon, Jordon, and Van Der~Schaar}]{yoon2018ganite}
Jinsung Yoon, James Jordon, and Mihaela Van Der~Schaar. 2018.
\newblock Ganite: Estimation of individualized treatment effects using
  generative adversarial nets.
\newblock In \emph{International Conference on Learning Representations}.

\bibitem[{Yuan et~al.(2019)Yuan, Hsia, Yang, Zhu, Chang, Dong, and
  Lin}]{yuan2019improving}
Bowen Yuan, Jui-Yang Hsia, Meng-Yuan Yang, Hong Zhu, Chih-Yao Chang, Zhenhua
  Dong, and Chih-Jen Lin. 2019.
\newblock Improving ad click prediction by considering non-displayed events.
\newblock In \emph{Proceedings of the 28th ACM International Conference on
  Information and Knowledge Management}, pages 329--338.

\bibitem[{Zhang et~al.(2021)Zhang, Feng, He, Wei, Song, Ling, and
  Zhang}]{zhang2021causal}
Yang Zhang, Fuli Feng, Xiangnan He, Tianxin Wei, Chonggang Song, Guohui Ling,
  and Yongdong Zhang. 2021.
\newblock Causal intervention for leveraging popularity bias in recommendation.
\newblock In \emph{Proceedings of the 44th International ACM SIGIR Conference
  on Research and Development in Information Retrieval}, pages 11--20.

\end{thebibliography}
\bibliographystyle{acl_natbib}

\appendix

\section{The detail of baselines}
\label{sec:appendix}
We compare our proposed approach with the following state-of-the-art dialogue classification models and multi-task models, which mainly come from MHCH, SSA and other similar tasks. We briefly categorize these baselines and introduce them below.

\textbf{Baselines for the MHCH task. }\textbf{HRN} \cite{lin2015hierarchical}: It uses a bidirectional LSTM to encode utterances and then fed these utterance features into a standard LSTM for context representation. \textbf{HAN} \cite{yang2016hierarchical}: HAN is a hierarchical network with two levels of attention mechanisms on word-level and utterance-level. \textbf{BERT} \cite{devlin2018bert}: It uses a pre-trained BERT model to construct the single utterance representations for classification. \textbf{HEC} \cite{kumar2018dialogue}: It builds a hierarchical recurrent neural network using bidirectional LSTM as a base unit and the conditional random field (CRF) as the top layer to classify each utterance into its corresponding dialogue act. \textbf{CRF-ASN} \cite{chen2018dialogue}: It extends the structured attention network to the linear-chain conditional random field layer, which takes both contextual utterances and corresponding dialogue acts into account. \textbf{HBLSTM-CRF} \cite{kumar2018dialogue}: It is a hierarchical recurrent neural network using bidirectional LSTM as a base unit and two projection layers to combine utterances and contextual information. \textbf{DialogueRNN}  \cite{majumder2019dialoguernn}: It is a method based on RNNs that keeps track of the individual party states throughout the conversation and uses the information for emotion classification. \textbf{CASA} \cite{raheja2019dialogue}: It leverages the effectiveness of a context-awar self-attention mechanism to capture utterance level semantic text representations on prior hierarchical recurrent neural network. \textbf{LSTMLCA} \cite{dai2020local}: It is a hierarchical model based on the revised self-attention to capture intra-sentence and inter-sentence information. \textbf{CESTa} \cite{wang2020contextualized}: It employs LSTM and Transformer to encode context and leverages a CRF layer to learn the emotional consistency in the conversation. \textbf{DAMI} \cite{liu2021time}: It utilizes difficulty-assisted encoding to enhance the representations of utterances, and a matching inference mechanism is introduced to capture the contextual matching features. 

\textbf{Multi-task baselines. }\textbf{MT-ES} \cite{ma2018detect}: It proposes a joint framework that unifies the two highly pertinent tasks. \textbf{JointBiLSTM} \cite{bodigutla2020joint}: It minimizes an adaptive multi-task loss function in order to jointly predict turn-level Response Quality labels provided by experts and explicit dialogue-level ratings provided by end users. \textbf{DCR-Net} \cite{2020DCR}: It considers the cross-impact and model the interaction between the two tasks by introducing a co-interactive relation layer.
\textbf{RSSN} \cite{liu2021role}: It integrates both dialogue satisfaction estimation and handoff prediction in one multi-task learning framework.


\end{document}